# Contrastive Learning for Knowledge-Based Question Generation in Large Language Models

1st Zhenhong Zhang
George Washington University
Washington, USA

2nd Jiajing Chen
New York University
New York, USA

3rd Weiyan Shi
Singapore University of Technology and Design
Singapore, Singapore

4th Lingjie Yi
Stony Brook University
Stony Brook, USA

5th Chihang Wang
New York University
New York, USA

6th Qian Yu*
Trine University
Detroit, USA

*Abstract*—With the rapid development of artificial intelligence technology, especially the increasingly widespread application of question-and-answer systems, high-quality question generation has become a key component in supporting the development of these systems. This article focuses on knowledge-based question generation technology, which aims to enable computers to simulate the human questioning process based on understanding specific texts or knowledge bases. In light of the issues of hallucination and knowledge gaps present in large-scale language models when applied to knowledge-intensive tasks, this paper proposes an enhanced question generation method that incorporates contrastive learning. This method utilizes multiple models to jointly mine domain knowledge and uses contrastive learning to guide the model in reducing noise and hallucinations in generation. Experimental results show that by designing prompts containing contrasting examples, the model's performance in question generation improves considerably, particularly when contrasting instructions and examples are used simultaneously, leading to the highest quality of generated questions and improved accuracy. These results demonstrate that the method proposed in this study, which combines contrasting context and chain-of-thought prompts, can effectively improve both the quality and the practicality of question generation.

## I. INTRODUCTION

The goal of question generation is to enable computers to simulate the human questioning process based on understanding specific text or knowledge base. In the field of question generation, knowledge base-based question generation is a key branch, which generates corresponding natural language questions by utilizing query subgraphs and answers in knowledge graphs [1].

Question generation tasks provide data support for question-answering systems. Question-answering system research requires high-quality question-answering data as training and test corpora [2]. As the scale of question-answering systems grows, the question-answering datasets that were previously only manually annotated can no longer meet the requirements of training and testing. With the accelerated development of the digital age, the application of artificial intelligence technology, especially in the field of question-answering systems, has become more and more extensive. Professional question-answering systems can provide accurate answers to complex queries and support decision-making and learning processes in multiple fields such as education, medical care, and law. To achieve this goal, the system needs to rely on a large amount of high-quality question-answering data as training and test corpora. However, as the scale of question-answering systems expands and domain knowledge deepens, the traditional method of manually annotating question-answering datasets can no longer meet current needs in terms of cost, efficiency, and coverage. With the in-depth application of question-answering systems in various fields, how to build high-quality question-answering datasets at low cost has become a research focus to meet the growing demand for data in these systems[3].

The emergence of large language models (LLMs), such as OpenAI's ChatGPT, marks a major shift in the field of artificial intelligence[4]. These models have demonstrated unprecedented capabilities in reasoning and generation tasks and have become the focus of the current artificial intelligence field. LLMs have mastered a wealth of language knowledge and world knowledge through pre-training on large-scale corpora. Researchers have found that the use of LLMs to synthesize data can already achieve the quality of manually annotated data [5]. By using large language models to generate domain questions, not only can more natural, richer, and more diverse questions be generated, but also complex domain knowledge can be understood and processed, providing users with more accurate and in-depth answers. The innovations of this paper are as follows:

A method for enhancing question generation based on contrastive context and thought chain is proposed. Based on the idea of contrastive learning, this paper first designs contrast prompts containing positive and negative examples to guide multiple a large models to generate candidate questions in parallel. Secondly, in order to screen the candidate questions, a thought chain prompt large language model is designed to score and evaluate the candidate questions to obtain high-quality questions.

## II. RELATED WORK

The task of question generation using large language models (LLMs) has made significant progress, particularly in knowledge-intensive domains. Various approaches leveraging deep learning techniques, such as generative adversarial networks (GANs) and neural networks, have been explored to enhance the accuracy and quality of generated questions. Large language models have revolutionized natural language processing (NLP) tasks, including question generation, by utilizing pre-trained models on vast corpora. Recent studies have applied advanced neural network techniques to language-related tasks, such as emotional analysis, demonstrating parallels with question generation tasks that require complex understanding and synthesis of information [6]. Research focusing on named entity recognition using advanced pre-trained models has provided valuable insights into how LLMs handle domain-specific knowledge, which is critical for knowledge-based question generation [7]. Additionally, work on optimizing text classification with graph neural networks has demonstrated how these techniques can improve LLM-driven tasks, particularly in terms of feature extraction and classification accuracy [8-10].

In recent years, contrastive learning has emerged as a powerful tool for enhancing model performance by guiding networks through positive and negative examples, a core technique in this study. Although traditionally more prominent in image-related tasks, contrastive learning has shown great potential in NLP. For instance, the use of ELMo word embeddings and multimodal transformers has exemplified how advanced NLP techniques can be integrated to improve language understanding and model performance, which aligns closely with the objectives of improving question generation through contrastive learning [11]. Furthermore, research involving GANs for image recognition has demonstrated the versatility of contrastive learning across different domains, which further supports its applicability in refining question-generation models [12-14].

To handle large-scale, knowledge-intensive tasks, optimizing neural networks is crucial for reducing hallucinations and addressing knowledge gaps. Studies in anomaly detection and risk assessment, particularly in financial markets, have applied deep neural networks to data-driven predictions [15], sharing the goal of reducing noise and enhancing model output accuracy, which is a key challenge in question generation tasks as well [16]. Moreover, techniques such as adaptive friction in deep learning optimizers, which utilize activation functions like sigmoid and tanh, have been introduced to improve model performance, contributing to the stability and precision of models designed for complex tasks like question generation [17].

Although the focus of this research is improving question generation, similar methodologies have been successfully applied in other domains, providing valuable insights for this study. For instance, research on dynamic hypergraph-enhanced prediction in medical data and the use of temporal dynamic graphs for fraud detection showcase the adaptability of advanced neural network methods, which are relevant to optimizing LLMs for knowledge-based tasks [18, 19]. These examples demonstrate the broader applicability of deep learning models across diverse, knowledge-intensive fields, offering foundational methods that can be applied to enhance question generation through contrastive learning.

Finally, the success of large-scale question generation relies not only on the architecture of models but also on the optimization of training algorithms. Studies on optimized gradient descent for neural network training have highlighted its importance for fine-tuning models in tasks such as question generation [20]. Additionally, the enhancement of convolutional neural networks through higher-order numerical methods and the development of hybrid frameworks [21], such as the LSTM-GARCH model, underline the need for continuous optimization in training to handle the complexity of large datasets and tasks [22, 23]. These advancements provide essential techniques that support the development of more efficient and accurate question-generation systems.

In conclusion, this paper builds upon the advancements in large language models, contrastive learning, and deep neural network optimization. By integrating contrastive learning with thought chain prompts, the proposed methodology addresses the limitations of current question-generation systems, such as hallucinations and knowledge gaps, especially in knowledge-intensive tasks.

## III. METHOD

Large language models have demonstrated excellent performance in a variety of natural language processing tasks by pre-training on a wide range of corpora. However, when faced with knowledge-intensive tasks, these models sometimes exhibit hallucinations and knowledge gaps due to the gap between model knowledge and domain knowledge. There is a lot of noise in the data generated by large models, and if these noisy data are used for training, they may harm the overall performance of the model. In addition, the diversity and complexity of natural language make traditional relevance indicators, such as ROUGE, insufficient to accurately reflect the true relevance of generated questions. Although manual inspection can provide accurate evaluation, it is costly and difficult to apply on a large scale. Therefore, how to effectively detect the quality of question generation and ensure the reliability and practicality of generated data has become an urgent problem to be solved.

In response to the above problems, this chapter studies the enhanced question generation method based on contrastive context and thought chain. For low-resource knowledge-intensive scenarios, large models are used for question generation. Domain knowledge is mined by combining multiple large models. At the same time, the idea of contrastive learning is adopted to prompt the large model to learn in a biased manner by selecting positive and negative examples to prevent noise and hallucinations. The thinking chain prompting large model is used to analyze the generated enhanced questions, select high-quality question samples, and realize the automated evaluation of the question generation process.

Input question dataset $D = \{<q, prgm, ans>\}$ to generate $k$ questions with similar semantics for each question $q$, and finally obtain the enhanced question dataset $Q^* = \{(q_i, q_{i,1}, q_{3,2}, ... q_{i,k})\}_{i=1}^{N}$, where $q_{i,k}$ is the Gth question generated based on question $k$.

The idea of contrastive learning is widely used in the field of deep learning. When distinguishing different types of data, the model learns the deep features of the data. Based on the idea of contrastive learning, a prompt text containing contrast examples is designed. The prompt text consists of three parts: task instructions, contrast context (positive examples and negative examples), and input, as shown in the formula:

$$prompt = \{Instruction, demo_{pos}, demo_{neg}, input\}$$

The task instructions first describe the task requirements for question rewriting, which is to rewrite the input question in the form of a question. The generated questions are of higher quality than the input questions, with richer language and more suitable for human users. Secondly, the model is required to analyze the differences between positive and negative samples and generate questions with quality closer to positive samples. The overall idea is as follows:

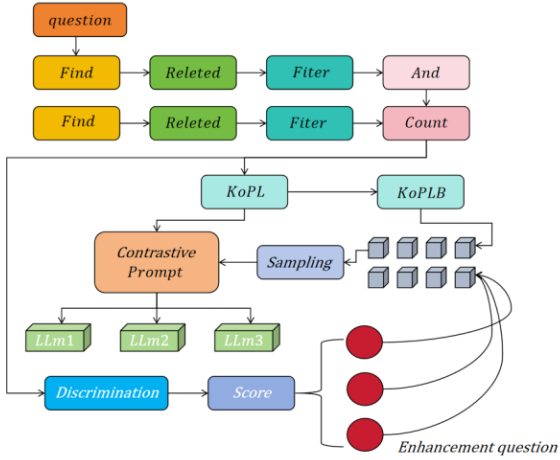

Figure 1 Overall design concept

The design idea of this paper is to first extract the logical form skeleton of the input question, search in two question pools, and select $m$ examples from them. They include positive example $demo_{pos}$ and negative example $demo_{neg}$. Each type of example contains $m$ examples. The search similarity calculation uses the overlap ratio of the bigrams, and the calculation is shown in the formula:

$$score = \frac{Count_{sample}(functions_{input})}{Count_{sample}(functions_{sample})}$$

The joint generations of multiple large models enriches the form of questions, but also leads to unstable question quality. In order to evaluate the quality of questions generated by different models, a scoring method based on the thinking chain is used to evaluate the questions, screen the questions generated in the middle, and select questions of different quality as positive and negative examples for subsequent generation. Using a large model different from the generation stage as a scoring model avoids the model being biased towards the results generated by itself.

For the enhanced question $Q = \{q_1, q_2, ..., q_n\}$ generated by the same logical form, the design thinking chain prompts the current question, guides the big model to make a brief evaluation of the question quality and semantic consistency, and gives a score. Define the scorer $f(\cdot)$, as shown in the formula:

$$Q^* = f(Q)$$
$$Q^* = \{(q_i, s_i, w_i)\}_{i=1}^{N}, \quad 1 \leq s_i \leq 5$$

The scorer evaluates the output questions of all large models to determine the relevance of the generated enhanced questions to the input questions and the corresponding logical forms. As shown in the formula, for an enhanced question $q$, if its question quality score and semantic consistency score are both greater than 3, the question will be retained and stored in the positive example question pool as a positive sample, otherwise it will be sent to the negative example question pool.

$$q = \begin{cases} Postive, & if \ s > 3 \ and \ w > 3 \\ Negtive, & otherwise \end{cases}$$

IV. EXPERIMENT

A. Experimental setup

The following large models are mainly used in this method. To keep scores independent, GLM-3-Turbo is used as the scoring model and the remaining models are generated as jointly. Use MBART as a test model.
(1) Llama3. Llama 3 is based on Llama 2 by expanding the vocabulary and using more corpus for further pre-training, which significantly improves Llama 2's understanding and generation capabilities.
(2) PaLM. PaLM (Pathways Language Model) is a large-scale language model developed by Google. It has powerful natural language processing capabilities and can perform tasks such as multi-round dialogue and text generation.
(3) Gopher. Gopher is one of the large-scale language models developed by DeepMind. It has a large scale and a large number of parameters and can achieve excellent results on a variety of language tasks.
(4) GLM-3-Turbo. GLM-3 is a bilingual pre-training model jointly released by the KEG Laboratory. Glm-3-Turbo performs a series of optimizations on the basis of GLM-3 to improve the generation speed.
(5) MBART. MBART is a sequence-to-sequence model pre-trained for multi-language denoising based on BART.

## B. Datasets

TriviaQA is a dataset designed to test the machine's common sense understanding and reasoning ability. It contains a large number of real-world question-answer pairs collected from books, movies, TV shows, and other sources. The characteristic of this dataset is that questions often need to span multiple documents to find answers, so it places high demands on the model's understanding and reasoning abilities. The TriviaQA dataset is divided into two versions: The web version and the Wikipedia version. The Web version is a question-answer pair collected from the Internet, while the Wikipedia version is a question-answer pair extracted from Wikipedia pages. Each question is associated with one or more documents, which can be web pages, Wikipedia pages, or other text resources to support the answer to the question. Its question-answer pair examples are shown in the following Table 1

Table 1 Dataset Example

| TriviaQA | Question | Answer |
| --- | --- | --- |
| 1 | Who was the first US president to resign? | Richard Nixon |
| 2 | What is the capital of Australia? | Canberra |
| 3 | Which character did Alan Rickman portray in the Harry Potter film series? | Severus Snape |

## C. Experimental Results

In order to verify that the method can be used as a means of data enhancement for the generation of question-answering data and improve the performance of the model, the experiment randomly selected 25% of the data for data enhancement and generated corresponding enhanced questions for each question. The effectiveness of data enhancement was verified by comparing the effects of different data enhancement methods on the accuracy of the model. The data enhancement experiment results are shown in Table 2

Table 2 Experimental Results

| Model | Acc |
| --- | --- |
| GLM-3.5 | 70.23 |
| GPT-3.5 | 67.12 |
| QKG-COT | 71.39 |
| Ours | |
| MBART-base | 72.31 |
| MBART-Repeat | 72.30 |
| MBART-Aug(ours) | 74.61 |

In order to investigate the impact of different prompt designs on the quality of question generation, the experiment used three settings to construct prompts. GPT-3.5, Llama3, Gopher, PaLM, and GPT3 were used for joint generation. Each model generated 3 enhanced questions. The prompt designs were as follows: Comparison instructions only: For each generated sample, there are only task instructions, requiring the model to generate better questions, but no examples of questions of different qualities are provided for the model to learn. Comparison examples only: For each generated sample, a positive example and a negative example are provided. Better questions and worse questions are marked in the system message, but there is no task instruction description. The model needs to learn the potential generation tendency from the examples. Comparison instructions and examples: For each generated sample, complete question generation instructions and better questions and worse questions are provided, and the model is required to learn example information and generate better questions for the question input. Table 3 shows the quality of questions generated using different prompt designs.

Table 3 The Experiment of Prompting Result

| Setting | positive | Negative | Acc |
| --- | --- | --- | --- |
| Only compare instructions | 471 | 513 | 47.87 |
| Comparison samples only | 832 | 203 | 80.38 |
| Comparing instructions and examples | 1032 | 13 | 98.80 |

The effect of generating only by comparing examples is better than that of generating only by comparing instructions, which shows that by providing examples, the model can learn the grammatical rules of logical form to a certain extent and generate them. When using both comparison instructions and examples, 934 of the generated questions are of qualified quality and 26 are of unqualified quality, and the accuracy rate is further improved to 98.8%. This shows that the strategy of combining comparison instructions and examples is the most effective in generating high-quality questions.

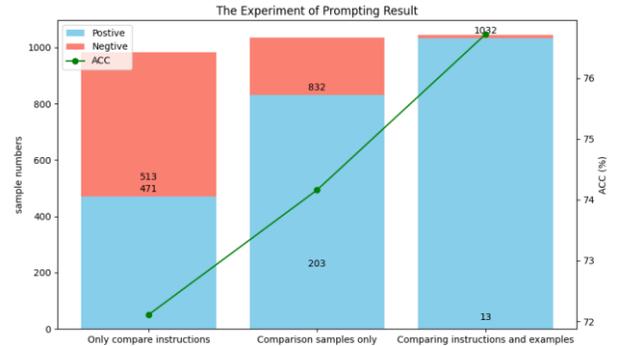

Figure 2 Experimental results bar graph

In order to display our experimental results more intuitively, we use a bar chart to show the above experimental conclusions. The results are shown in Figure 2.

## V. CONCLUSION

This paper explores a large-scale language model-enhanced question generation method based on contrastive context and thought chains. To address the problems of hallucinations and knowledge gaps that may occur in models in knowledge-intensive tasks, we propose a method that combines large-scale models to mine domain knowledge and introduce the idea of contrastive learning to guide large-scale models to perform bias learning by selecting positive examples and counterexamples, thereby reducing noise and hallucination phenomena. At the same time, the thinking chain is used to prompt questions generated by large-scale model analysis, screen high-quality question samples, and realize automated evaluation of the question generation process. Experimental results show that this method has significantly improved the quality of question generation, indicating that combining contrasting context and thinking chain prompts can effectively improve the quality and reliability of question generation, and is of great value in building a high-precision automatic question generation system. In addition, the study also verified the importance of contrasting examples. When contrasting instructions and examples are used at the same time, the quality of question generation reaches the optimal level, and the accuracy rate is further improved, proving the effectiveness of the strategy of combining contrasting instructions and examples. sex. The research in this article not only provides new ideas for solving the limitations of large-scale language models in knowledge-intensive tasks but also points out the direction for the development of future question-generation technology.